\documentclass[sigconf]{acmart}
\usepackage{subcaption}
\AtBeginDocument{%
  \providecommand\BibTeX{{%
    \normalfont B\kern-0.5em{\scshape i\kern-0.25em b}\kern-0.8em\TeX}}}

\setcopyright{none}
\renewcommand\footnotetextcopyrightpermission[1]{}
\usepackage{multirow}

\acmConference[UrbComp '20]{UrbComp '20: The 9th SIGKDD International Workshop on Urban Computing}{August 24, 2020}{San Diego, CA}

\usepackage[font=small,skip=1.25pt]{caption}
\usepackage[skip = 0.15pt]{subcaption} %

\begin{document}

\newcommand{\name}{{\tt AtCoR}}
\title{
Towards Dynamic Urban Bike Usage Prediction for Station Network Reconfiguration
}

\author{Xi Yang and Suining He}
\affiliation{%
  \institution{The University of Connecticut}
  \city{Storrs}
  \state{Connecticut}
}
\email{{xi.yang, suining.he}@uconn.edu}
\begin{abstract}

Bike sharing has become one of the major choices of transportation for residents in metropolitan cities worldwide.
A station-based bike sharing system is usually operated in the way that a user picks up a bike from one station, and drops it off at another. 
Bike stations are, however, not static, as the bike stations are often reconfigured 
to accommodate changing demands or city urbanization over time. 
One of the key operations is to evaluate candidate locations and install new stations to expand the bike sharing station network.
Conventional practices have been studied to predict existing station usage, while evaluating new stations is highly challenging due to the lack of the historical bike usage. 

To fill this gap, in this work we propose a novel and efficient bike station-level prediction algorithm called \name{}, which can predict the bike usage at
both existing and new stations (candidate locations during reconfiguration). 
In order to address the lack of historical data issues, virtual historical usage of new stations 
is generated according to their correlations with the surrounding existing stations, for \name{} model initialization. 
We have designed novel station-centered heatmaps which characterize for 
each target station centered at the heatmap the trend that riders travel 
between it and the station's neighboring regions, enabling the model to capture the learnable features
of the bike station network. 
The captured features are further applied to 
the prediction of bike usage for new stations. 
Our extensive experiment study on more than 23 million trips from three major bike sharing systems in US, including New York City, Chicago and Los Angeles, shows that \name{} outperforms baselines and state-of-art models in prediction of both existing and future stations.

\end{abstract}

\keywords{bike sharing, usage prediction, station-level, new stations, 
pick-ups and drop-offs,
attention, spatio-temporal
}
\maketitle

\section{Introduction}
Thanks to mobile networking and location-based services, bike sharing has become one of the major  transportation modalities for urban residents worldwide due to its convenience and efficiency.
As a representative product of the sharing economy, it is often hailed as the excellent helper
to solve the ``last mile'' problem in citizen transportation. 
Given the social and business importance, the bike sharing market is estimated to hit US\$5 billion by 2025\footnote{https://www.globenewswire.com/news-release/2019/12/03/1955257/0/en/Bike-Sharing-Market-is-Predicted-to-Hit-5-Billion-by-2025-P-S-Intelligence.html}.

\begin{figure}
    \centering
    \includegraphics[width=\columnwidth]{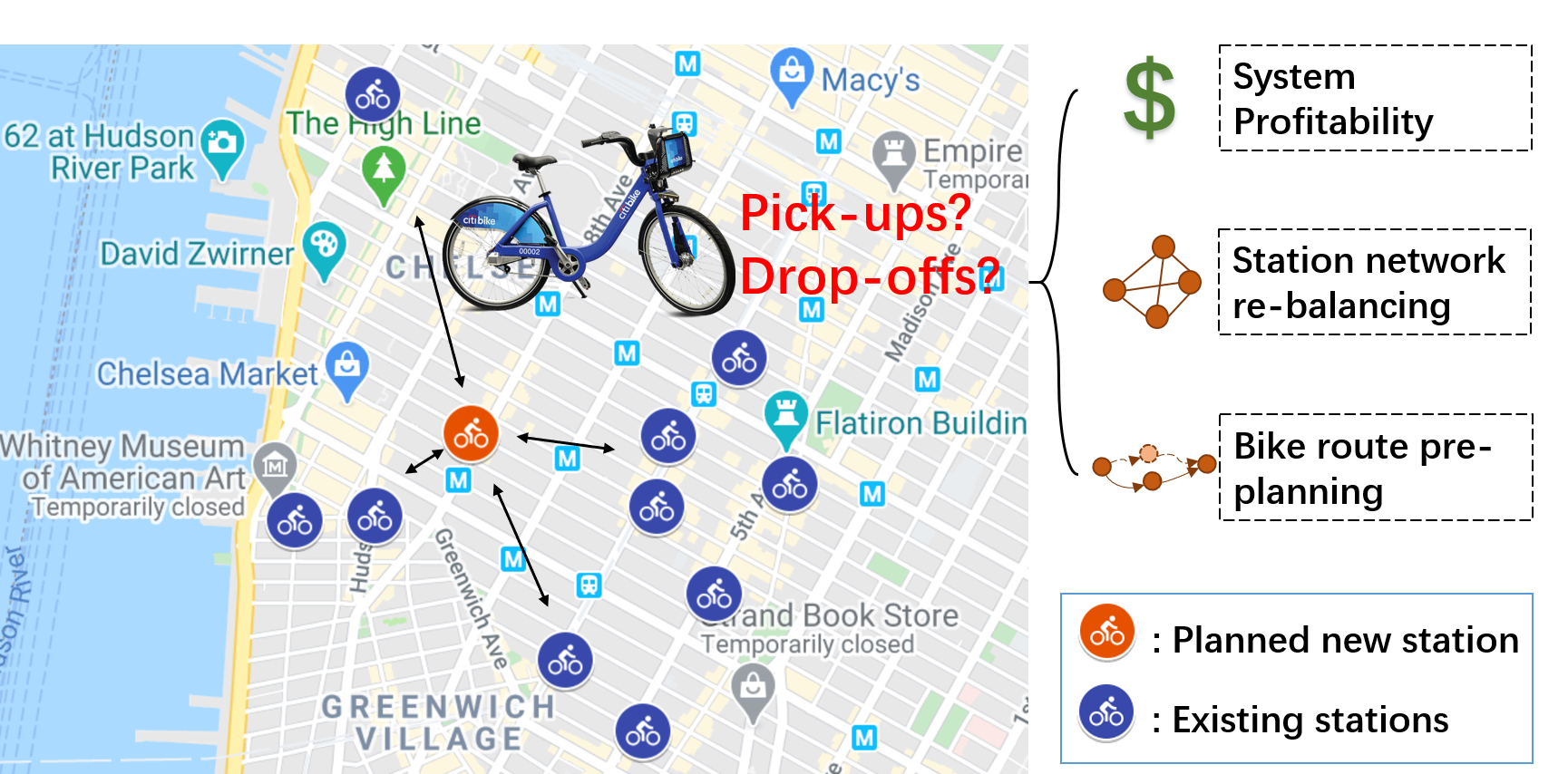}
    \caption{Illustration of bike sharing system and new station usage prediction.}
    \label{fig:problem}
    \vspace{-0.2in}
\end{figure}

A station-based bike sharing system (each station is equipped with multiple docks for bike parking) is usually operated in the way that a user or rider picks up a bike from one station (pick-ups) and drops it off at another (drop-offs). 
All the resultant bike trips (usage that consists of pairs of pick-ups and drop-offs) connect different parts of the city, forming \textit{the bike sharing station network}. 
The bike stations are, however, not static, as the bike sharing operators
often \textit{reconfigure} the stations to accommodate changing demands or city urbanization
over time \cite{he2018re, liu2015station}, where a key operation is to evaluate candidate locations and then install new stations to expand the bike sharing station network \cite{zhang2016bicycle,liu2017functional} as shown in Figure \ref{fig:problem}. 
It is essential
for the operators 
to know the potential bike usage of a future station at a certain candidate location beforehand,
which helps predicting the station profitability as well as 
its positive/negative effect upon the local mobility and traffic networks, 
enhancing the bike sharing service quality to the local community. 
This further benefits the future bike sharing network operations, including station re-balancing \cite{o2015data,liu2016rebalancing,hulot2018towards,wang2018bravo, pan2019deep}
and bike route pre-planning \cite{zhang2016trip}.

Despite the business and social importance, 
predicting bike usage at future stations is extremely challenging due to the lack of historical bike usage data of those stations. Conventional practices include user survey, crowdsourced feedbacks and public hearing, investigating the local demands, which is often costly and time-consuming.
Operating regulations are also provided for bike sharing operators as guidance on candidate station locations, which is based primarily on geo-information such as distance to road  intersections/public transit stations and station network density\footnote{\url{https://www.transformative-mobility.org/assets/publications/The-Bikeshare-Planning-Guide-ITDP-Datei.pdf}}. However, such guidance is often too general to take into account uniqueness of different cities, causing inefficiency in station reconfiguration as well as
reduced service quality and profit loss.
While most of the existing studies focus on predicting the bike usage at the existing stations, only a few of them have comprehensively explored how to forecast new stations for the station reconfiguration purposes.

\begin{figure}
    \centering
    \includegraphics[width=\columnwidth]{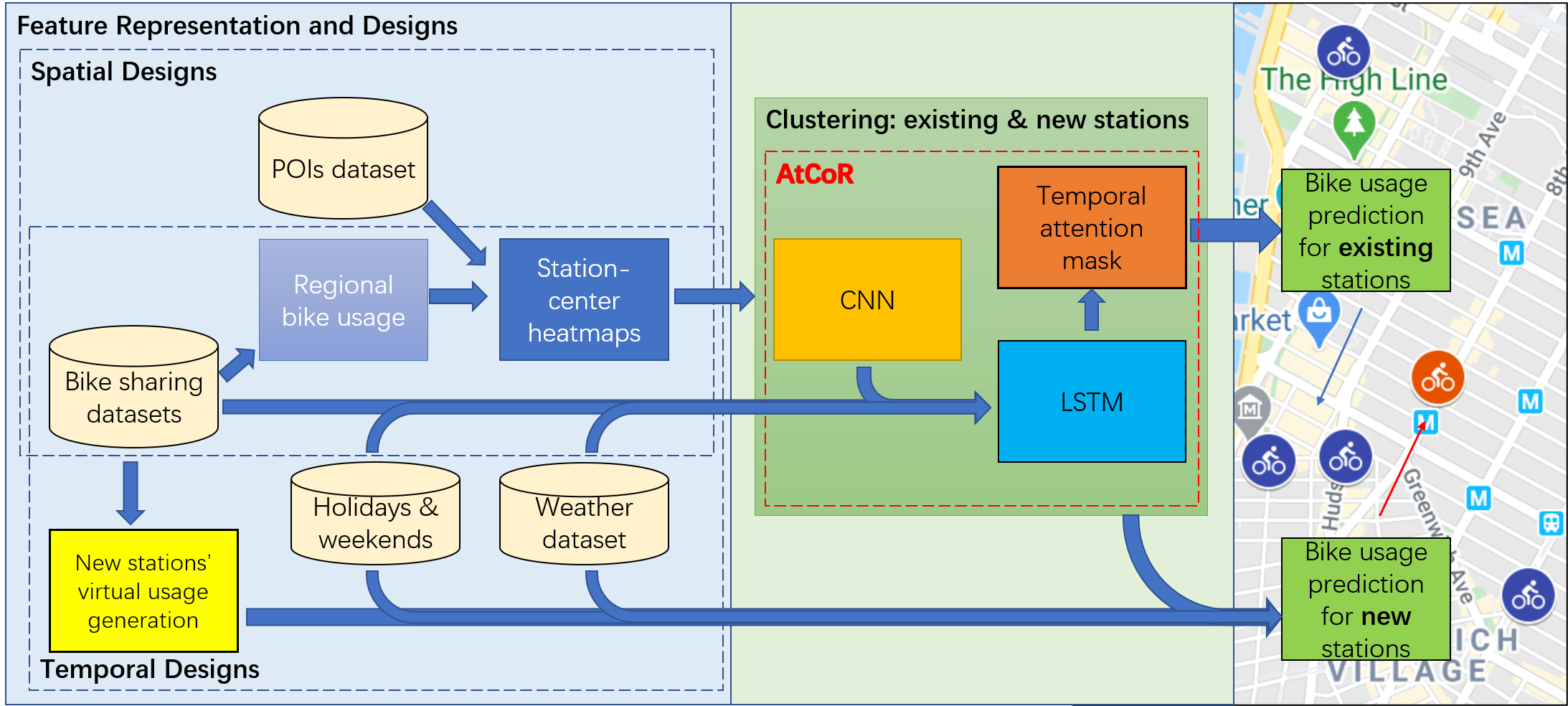}
    \caption{System overview and information flow of \name{}.} %
    \vspace{-0.2in}
    \label{fig:informationflow}
\end{figure}

To fill this gap,
in this work we propose a novel bike station prediction algorithm called \name{}, based on \underline{At}tention \underline{Co}nvolutional \underline{R}ecurrent Neural Network, for predicting station-based bike usage of future stations.
To tackle the lack of historical data issues for future/new stations to be deployed, 
virtual historical usage of new stations is generated according to their correlation with the surrounding existing stations. 
To predict the usage of new bike stations for reconfiguration decisions, 
we have designed novel station-centered
heatmaps which characterize for each target station centered at the heatmap
the trend that riders travel between it and the neighboring regions,
so that the model is able to capture the learnable common patterns of the bike station network through a convolutional neural network (\texttt{CNN}). A Long Short-Term Memory (\texttt{LSTM}) neural network with temporal attention mask leverages the common patterns with integration of historical data and external factors, such as weekends, holidays, and weather, to predict the bike usage for existing stations.  The captured bike usage features are then used to predict the pick-ups and drop-offs for new stations along with the virtual historical usage. 
An overview of information flow of this work is illustrated in Figure \ref{fig:informationflow}, summarizing the above process. %

Our main technical contributions are as follows:
\begin{itemize}
  \item[1)] \textit{Comprehensive bike data analysis for station 
  prediction designs}: We have conducted comprehensive and detailed real-world data analysis on how the weather conditions, regional bike usage and surrounding POIs impact the bike usage in the metropolitan area like New York City (NYC), Chicago and Los Angeles (LA), and visualized them to validate our design insights for station reconfiguration prediction. These features serve as the shared patterns leveraged for the bike usage prediction, including predicting for the new stations.
  \item[2)] \textit{Dynamic urban bike usage prediction for reconfigured station networks}:
   We have
  designed a novel scheme called \name{} to predict the
  station-level bike usage for new/future stations (candidate locations for reconfiguration) as well as existing/fixed ones. 
  We propose a novel design of station-centered feature heatmap representation as an input of \name{}, 
  which calculates the differences between the features at the location of each station and those at the surrounding areas. 
  The heatmaps account for the trend that riders travel from the center station to the neighborhood. 
  The inclusion of the heatmaps significantly
  improves the accuracy of usage predictions. Heatmaps are then fed to \name{} which consists of a deep Convolutional Neural Network (\texttt{CNN}) and a Long Short-Term Memory (\texttt{LSTM}) network with integration
  of temporal attention mechanism. 
  \item[3)] \textit{Extensive experimental studies with real-world datasets}: We have conducted extensive experimental studies upon 23,955,989 
  trips in total, across three major bike sharing systems in the United States: 
  20,551,697 from Citi Bike in New York City (NYC),
  3,113,950 from Divvy in Chicago, 
  and 290,342 from Metro Bike in Los Angeles (LA). 
  Our experimental studies, upon both the existing and new stations in the bike station network reconfiguration, have validated that
  \name{} outperforms the other baseline models in multi-station predictions, often by more than 20\% error reduction. 
\end{itemize}

While our studies focus on station-based bike sharing, the model and algorithm derived can be further extended to other transportation platform with deployment and expansion operations, including ride/car sharing \cite{chen2017data,jauhri2017space,10.1145/3331450} and scooter sharing \cite{he2020dynamic}.

The rest of the paper is organized as follows. We first conduct a brief survey on related studies in Section \ref{sec:related}. Then we analyze the datasets, and find out the features and their corresponding representations for model input Section \ref{sec:data}. 
Then the model is further presented in Section \ref{sec:model}. 
Afterwards, we present our extensive experimental studies to evaluate the performance of \name{} for both existing and new stations in Section \ref{sec:experiment}.
We will discuss the deployment in Section \ref{sec:discussion}, and conclude in Section \ref{sec:conclusion}.

\section{Related Work}\label{sec:related}

Traffic flow prediction
has been studied recently due to the advances of intelligent transportation and smart cities. Gong et al. proposed a potential passenger flow predictor to help the decision on the places to construct new metro station \cite{gong2019potential}. 
Tang et al. tackled the dense and incomplete trajectories for citywide traffic volume inference \cite{tang2019joint}. For dockless bike sharing system, the potential bike distribution and the detection of parking hotspots in a new city has been made by Liu et al. \cite{liu2018inferring, liu2018will}. 
Different from above works on dockless bike sharing, we focus on station-based deployment
due to its wider social acceptance, and our studies provide highly 
comprehensive studies of predicting usage for new/future stations before the bike sharing station reconfiguration \cite{he2018re}. 

Based on bike usage prediction granularity, there are three categories of prediction models in current works in the bike sharing systems: \textit{city-level}, \textit{cluster-level}, and \textit{station-level} \cite{li2019learning}. In \textit{city-level} prediction, the aim is to predict bike usage for a whole city, while at the \textit{cluster} level the goal is to predict bike usage for clusters of bike stations. The station cluster is generated by clustering algorithms such as the Bipartite Station Clustering \cite{li2015traffic}, the Community Detection and the Agglomerative Hierarchical Clustering method \cite{zhou2015understanding}, the K-means clustering and the Latent Dirichlet Allocation (LDA). Chen et al. further considered clusters as dynamic rather than static \cite{chen2016dynamic}. While city-level and cluster-level predictions save the computational cost by simplifying the problems, station-level \cite{hulot2018towards,chai2018bike,chen2020predicting, he2020towards} prediction still benefits the bike sharing system management the most, including fine-grained station rebalancing,  but yet is challenging.

Recent research efforts have also been made upon
traffic prediction beyond the bike sharing systems. 
Deep learning approaches 
have been studied for traffic flow prediction \cite{lv2014traffic,huang2014deep, yu2017spatio, fadlullah2017state, he2020dynamic, he2019spatio}. 
Yao et al. proposed a \texttt{CNN-LSTM} based transfer learning method to predict traffic flows across different cities \cite{yao2019learning}. Pan et al. used a deep meta learning model to predict urban traffic \cite{pan2019urban}. Ma et al. constructed spatial-temporal graphs of urban traffic which was learned by a deep convolutional neural network \cite{ma2017learning}. 
In this work, we study a novel approach based on deep learning designs and data-driven studies
to handle the new stations usage prediction problems.
Specifically, we propose a novel approach \name{} which consists of a \texttt{CNN} component for modeling spatial characteristics and a \texttt{LSTM} with integration of a temporal attention mechanism for capturing temporal characteristics .

\section{Data Analysis \& Feature Designs}\label{sec:data}

Because no historical usage data of new stations is available, future mobility patterns might only be predicted through accessible spatial and temporal characteristics of the stations. In this work, we propose using station specific regional usage, points of interest (POIs), weather conditions and holidays as input features as they are all available once the locations and launching time of new stations are provided by system operators. 

In this section, we first present the preliminary concepts defined for the feature designs of bike sharing stations in Section \ref{sec3.1}. 
Then in Section \ref{sec3.3} we discuss our representation design of the features we choose, namely \textit{station-centered heatmaps}, which significantly improve the model performance. We further cluster station networks based on such heatmaps to save computational cost of training.

Table \ref{tab:symbol} summarizes all the symbols as well as their definitions in this work.

\begin{table}[]
\caption{List of symbols and definitions}
\small
\label{tab:symbol}
\begin{tabular}{l|l}
\hline
Symbols & Definitions \\ \hline
$n$ &  New stations \\
$f$ &  Existing stations \\
$\mathbf{F}$ & Set of existing stations\\
$j,k$ &  Index of the stations \\
$T, S$ &  Ranges of timestamps \\
$t,\tau$ & Indices of timestamps  \\
$\mathcal{T}$& Number of timestamps\\
$g_{lat}, g_{lon}$ & Height and width of each grid \\
$\mathcal{G}_{lat}, \mathcal{G}_{lon}, \mathcal{P}$ & Dimensions of station-centered heatmaps\\
$\mathbf{H}$ & Station-centered heatmaps\\
$e$ & Entries of station-centered heatmaps\\
$L$ & Station usage\\
$h, w, c$ & Parameters of the convolutional layers\\
$\mathbf{x}$ & Input of \texttt{LSTM}\\
$\mathbf{h}$ & \texttt{LSTM} hidden state\\
$\mathbf{c}$ & \texttt{LSTM} cell state\\
$d$ & Number of hidden units of \texttt{LSTM}\\
$\mathbf{v}, \mathbf{W},\mathbf{U}, \mathbf{b}$ & Trainable parameters\\
$\lambda, \gamma$ & Attention scores between a pair of states\\
$\mathbf{d}$ & Decoder's input\\
$\mathbf{ex}$ & External features\\
$\omega$ & \begin{tabular}[c]{@{}l@{}}Similarity scores between a new station and \\ its surrounding existing peers\\
	\hline
\end{tabular} 
\end{tabular}
\end{table}

\subsection{Preliminary for Station Feature Studies}\label{sec3.1}

\subsubsection{Overview of Datasets and New/Existing Stations}
We conduct our data analytics upon
three datasets: 
Citi Bike of NYC\footnote{\url{https://www.citibikenyc.com/}} of 2019 (20,551,697 trips), 
Divvy Bike of Chicago\footnote{\url{https://data.cityofchicago.org/Transportation/Divvy-Trips/fg6s-gzvg}} of the first three quarters of 2019 (3,113,950 trips), 
and Metro Bike of Los Angeles\footnote{\url{https://bikeshare.metro.net/}} of 2019 (3,113,950 trips).

In this study, the \textit{new stations} established in a set of certain time intervals $[T, T']$ are defined as the ones that have no historical bike usage data in the past 30 days, i.e. from $(T/24 - 30)$ days to $T$ considering $T$ on hourly basis, while the \textit{active existing stations} (or existing stations in short) in a certain time interval $[S, S']$ 
are defined as the ones that 
have historical usage data every day
in this time interval.
Note that active existing stations focus on the existing stations which are populous with everyday bike usage.

\begin{figure}
    \centering
    \includegraphics[width=\columnwidth]{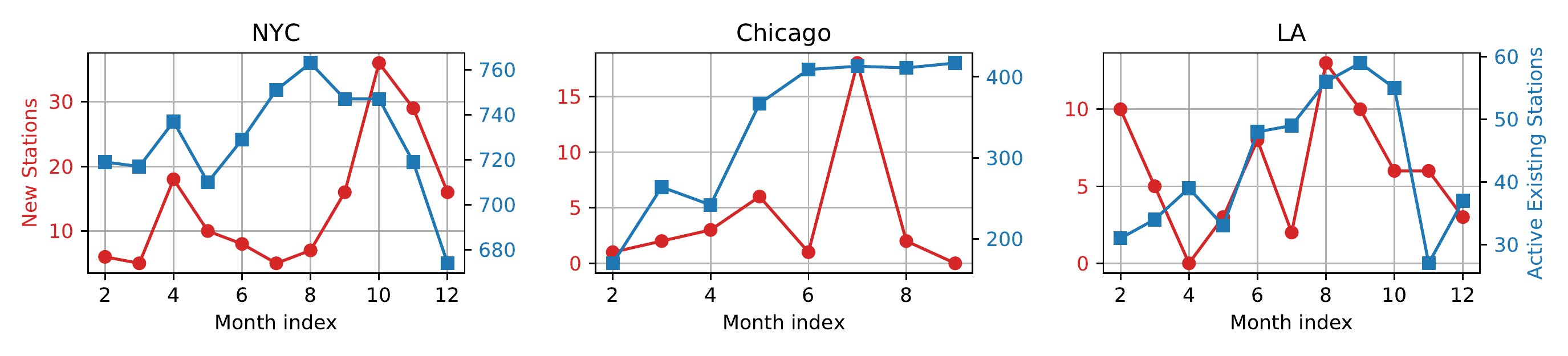}
    \caption{The number of active existing stations and new stations per month in NYC, Chicago and LA.
    }
    \label{fig:newfixed_stat}
\end{figure}

For example, 
there are 
454 active existing stations at the New York City (NYC) in 2019 in total, and from April 11, 2019 to July 19, 2019 there are 631 existing stations, while in June 2019 there are 8 new stations established, compared to totally 1,047 unique bike station coordinates in the whole year including those stations that are newly installed, removed, and relocated. The monthly number of active existing stations and new stations are shown in Figure \ref{fig:newfixed_stat} for the three datasets as demonstration.

We select the following spatial and temporal features in our modeling: 
regional usage in areas where bike stations are located, points of interest, weather conditions and holidays and weekends.
Since there is no historical bike usage at the future locations of the new bike stations, it is difficult to directly characterize the mobility patterns for the target locations.
However, the mobility patterns are strongly related to the spatial and temporal features of those locations. 
Thus, in this work \name{} incorporates the correlations with the above features for the new station prediction. 

\subsubsection{Regional Usage}

To characterize the urban bike usage in a computationally efficient manner, we discretize the neighborhood city map around each station into grids, each of which is an $g_{lat}$ m $\times$ $g_{lon}$ m rectangular region. 
The total bike pick-ups/drop-offs within the grid can represent the bike usage popularity of this specific region. Bikes rented at the popular regions tend to be returned at surrounding regions. 
Therefore, the station-centered regional usage distribution is chosen an essential input feature. 

\subsubsection{Points of Interest (POIs)}
Another key insight of feature selection is that the differences in the POI distributions
around each station can steer the bike riders' travels with the corresponding preferences or purposes. 
Therefore, the POI distributions
are used as another important features. Following the manners of defining regional station usage, the POI distributions
are defined as the total numbers of POIs within a $g_{lat}$ m $\times$ $g_{lon}$ m grid for each POI category.

\begin{figure}
    \centering
    \includegraphics[width=\columnwidth]{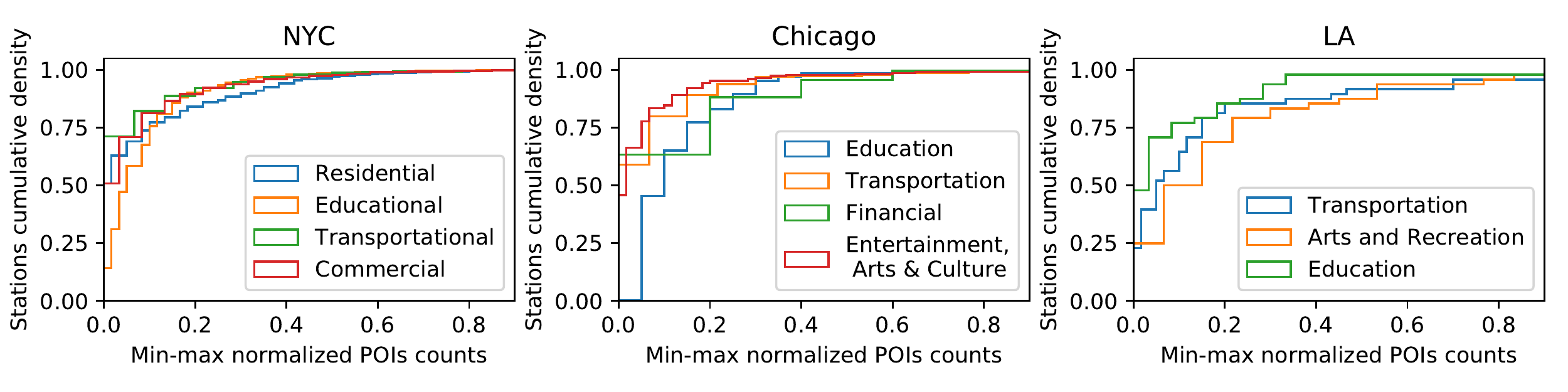}
    \caption{POIs distributions around bike stations in NYC, Chicago and LA.}
    \label{fig:poisdist}
\end{figure}

Figure \ref{fig:poisdist} illustrates some examples of the different distributions of different categories of POIs in a 500m $\times$ 500m grid region. POIs around active existing stations from April 11, 2019 to July 19, 2019 are shown here with the numbers of POIs normalized by the min-max method. 

Specifically, the POIs data of NYC are obtained through NYC Open Data\footnote{\url{https://data.cityofnewyork.us/City-Government/Points-Of-Interest/rxuy-2muj}} which contain 13 major categories, such as residential, education facility, cultural facility, recreational facility, social services, transportation facility, commercial, government facility (non public safety), religious institution, health services, public safety, water and others.
The POIs of Chicago are obtained through OpenStreetMap Overpass API\footnote{\url{https://wiki.openstreetmap.org/wiki/Overpass_API}} where POIs are categorized by the OSM tag of \texttt{amenity} including sustenance, education, transportation, financial, healthcare, entertainment, arts $\&$ culture, and others.
The POIs of LA are obtained through LAC Open Data\footnote{\url{https://data.lacounty.gov/}}
which include communications, transportation, private industry, health and mental health, social services, postal, arts and recreation, community groups, municipal services, public safety, education, government, emergency response, physical features and environment.

\subsubsection{Weather Conditions, Holidays and Weekends}
\begin{figure}
    \centering
    \includegraphics[width=\columnwidth]{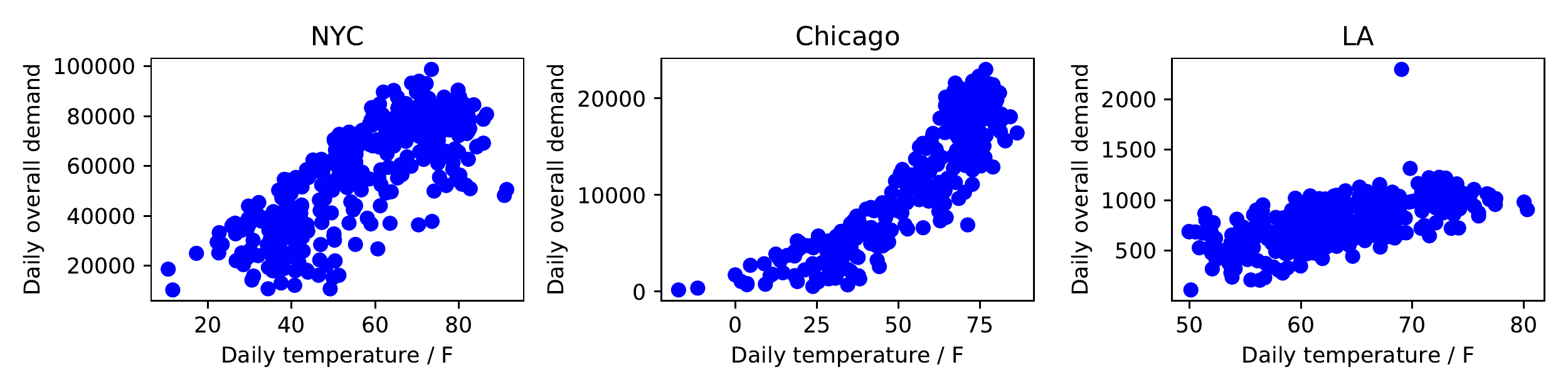}
    \caption{Temperature effect on daily overall bike usage.}
    \label{fig:temp}
\end{figure}

\begin{figure}
    \centering
    \includegraphics[width=\columnwidth]{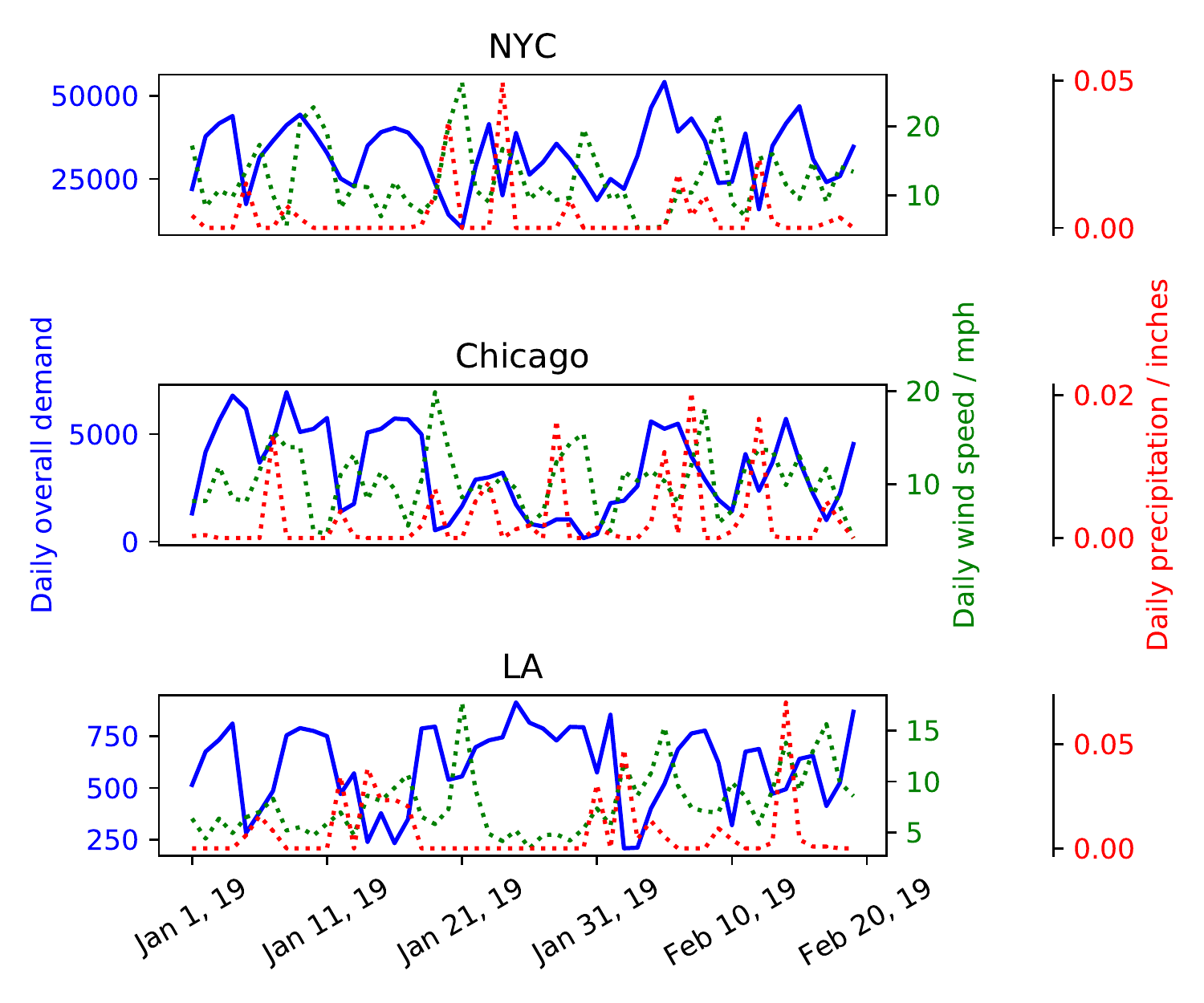}
    \caption{The effect of precipitation and wind speed on city's daily overall bike usage.}
    \label{fig:prec}
\end{figure}

The station usage is highly correlated with the weather conditions. 
We have collected and analyzed the weather condition data from 
open source weather data API\footnote{\url{https://api.weather.com}}.
We have analyzed the correlations between weather conditions and the station usage. 
Analysis of one-year bike usage
reveals that both high and low temperature decreases the bike usage. 
The effect of daily temperature, precipitation and wind speed on daily overall usage of the city is illustrated in Figures \ref{fig:temp} and \ref{fig:prec}. Clearly, 
precipitation, including rain, snow and large wind speed, significantly 
reduces the bike usage.
Given above, the weather conditions including temperature, precipitation 
and wind speed are chosen as the input features. 
Besides weather conditions, the bike usage has different patterns on federal holidays
and weekends from that on workdays. 
Specifically, we set the indicator as 1 if a time interval belongs to holiday/weekend periods, or 0 for weekdays otherwise.  
As an example, for the time interval of [0:00 a.m., 1:00 a.m.], 2019-01-01,
the external vector including temperature, wind speed, precipitation and holiday/weekends,
is given by [47 \textdegree{}F,  1.5 mph,  0.08 in,  1].

\subsection{Feature Representations \& Designs}\label{sec3.3}
Given the spatial and temporal features presented above for each station, we present a representation design
to integrate them as the model inputs, which will be discussed in details as followed.

\subsubsection{Station-centered Heatmaps}\label{heatmaps}
\begin{figure}
    \centering
    \begin{subfigure}[b]{0.225\textwidth}
         \centering
         \includegraphics[width=\textwidth]{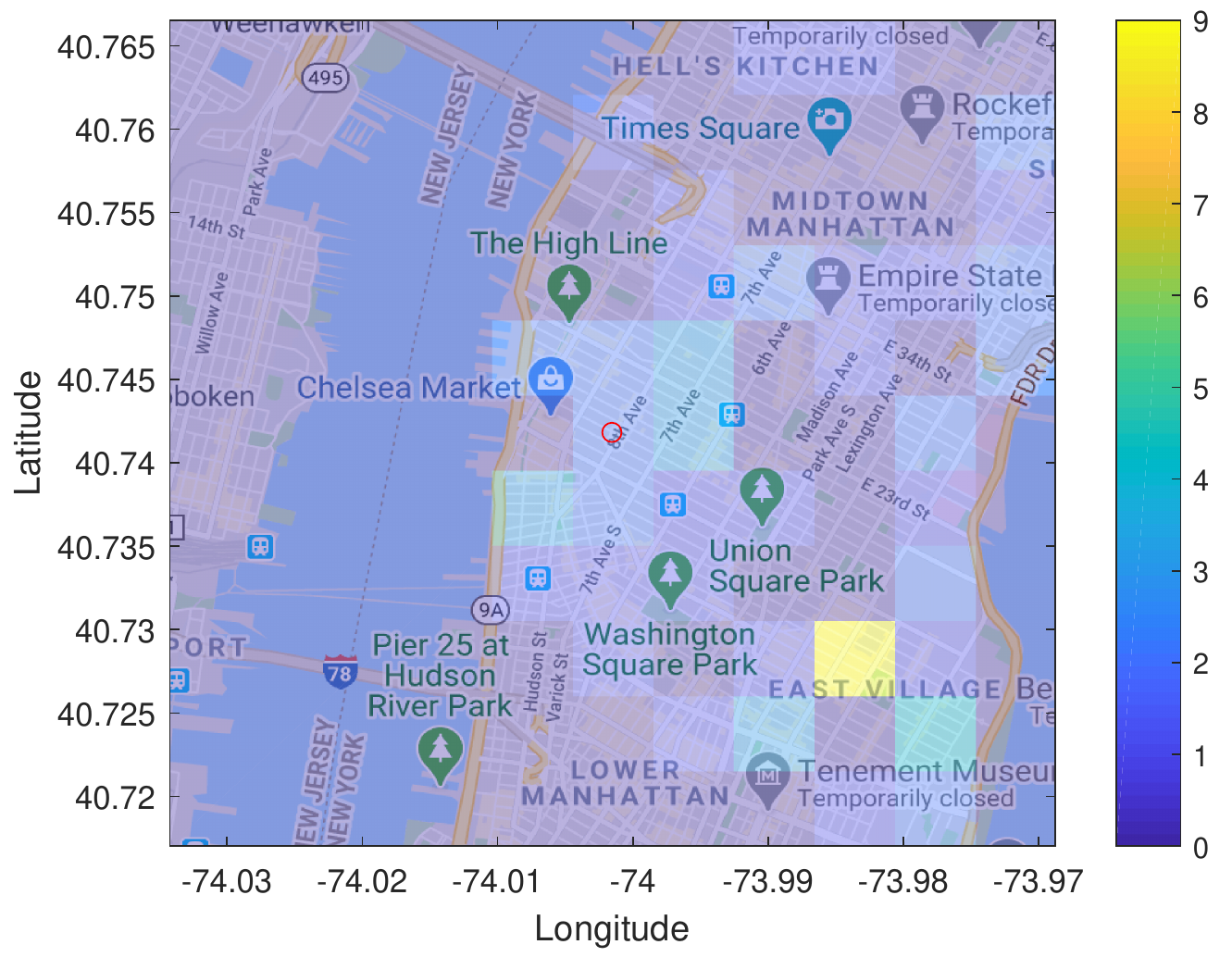}
         \caption{}
    \end{subfigure}
    \begin{subfigure}[b]{0.225\textwidth}
         \centering
         \includegraphics[width=\textwidth]{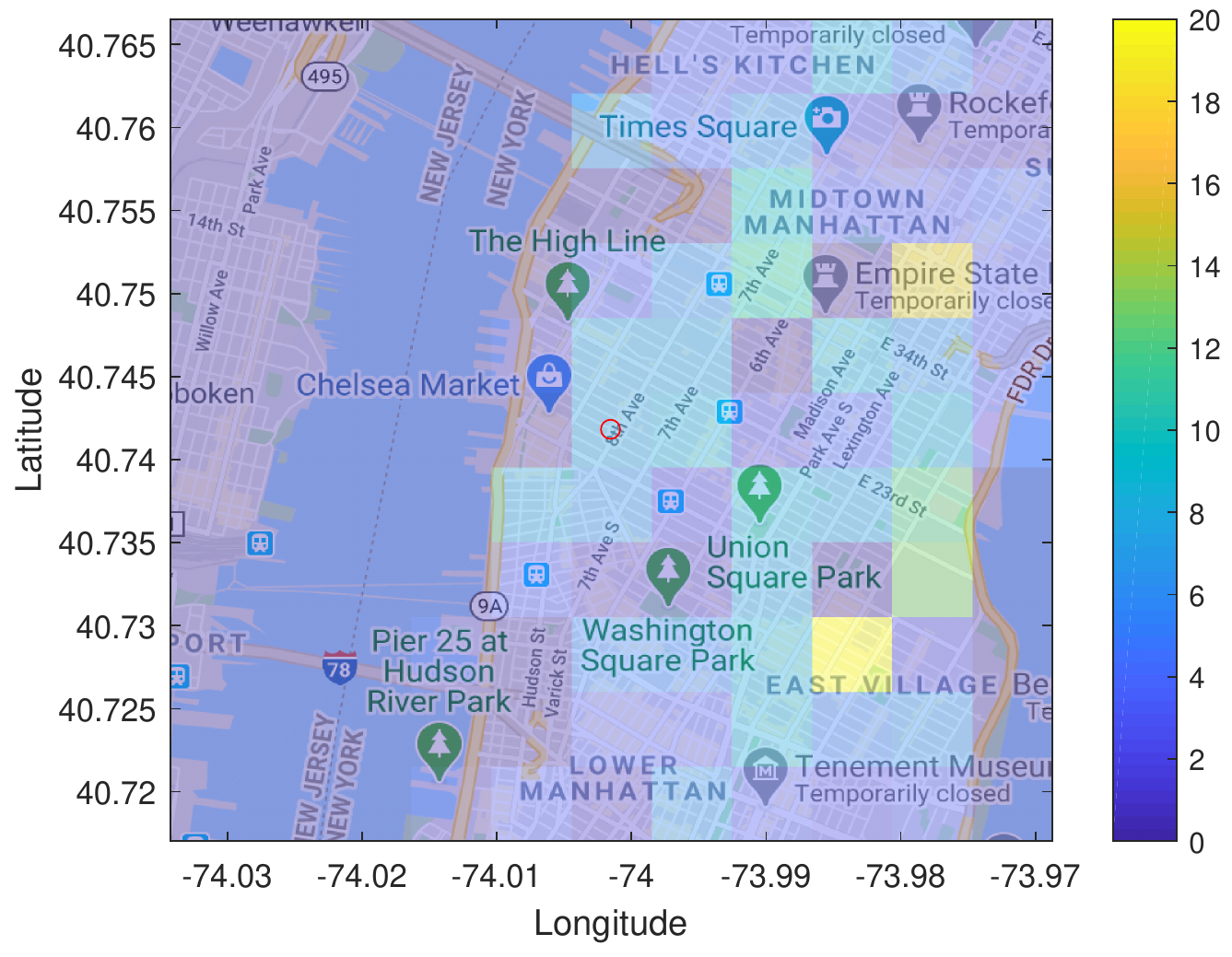}
         \caption{}
    \end{subfigure}
    \caption{Regional pick-ups for a candidate station (red cycle) in Manhattan, NYC on 
    (a) Sunday 01-20-2019 from 8:00 a.m. to 9:00 a.m.;
    (b) Monday 01-21-2019 from 8:00 a.m. to 9:00 a.m.}
    \label{fig:heatmap}
\end{figure}

We construct a $\mathcal{G}_{lat}\times\mathcal{G}_{lon}\times\mathcal{P}$ heatmap centered at the stations studied where each grid is a $g_{lat}$ m $\times$ $g_{lon}$ m area on city map with $\mathcal{P}$ channels including the regional usage and POIs distribution of the grid area. The first two channels of the heatmap are the regional pick-ups and drop-offs, respectively, with each of the following channel as the POIs distributions of one POIs category. Every entry of the heatmaps is the regional usage or the POIs amount for the corresponding grid location. 
Figure \ref{fig:heatmap} shows a $11\times11$ regional pick-ups as the first channel of the station-centered heatmap for a candidate station in NYC.

After construction of the heatmaps centered at a particular station, the heatmaps are normalized by subtracting each one of the $\mathcal{G}_{lat}\times\mathcal{G}_{lon}$ grid features by the features of the center grid. 
This way, 
the normalized heatmaps represent 
the riders' motivations or mobility trends departing from this station to the neighborhoods, characterizing the spatial-temporal features near a station. 

\subsubsection{Station Clustering}\label{seccluster}
To enhance the learning efficiency upon large-scale bike station network, 
we design a clustering scheme for 
the bike stations and train the 
\name{} for certain cluster. 
We cluster existing stations and newly established ones all together. 
This way, we can save computational cost by reducing the amount of stations being trained, while the predictions upon new stations can be enabled by the patterns learned from other stations.

\begin{figure}
    \centering
    \begin{subfigure}[b]{0.55\columnwidth}
         \centering
         \includegraphics[width=\textwidth]{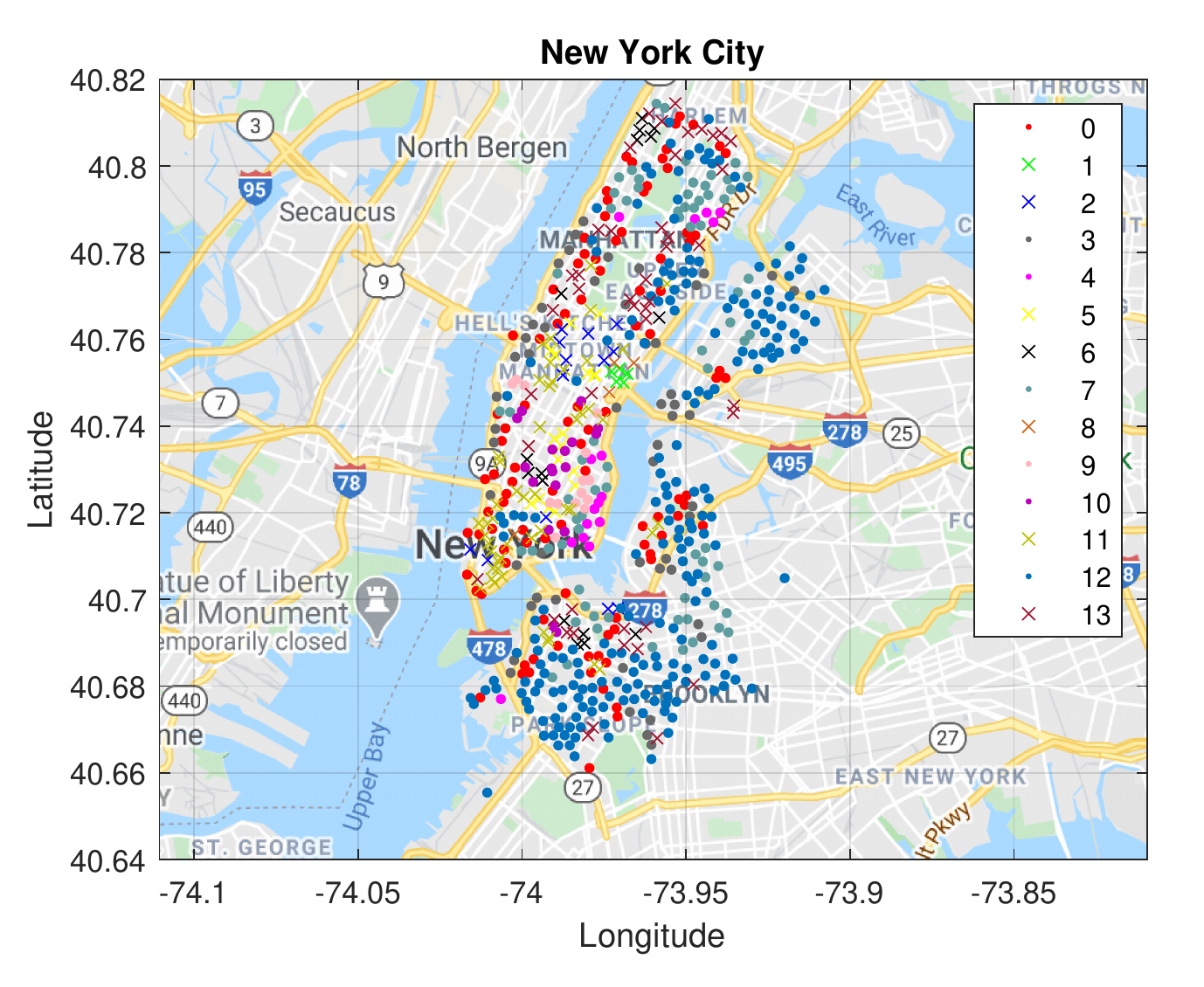}
         \label{fig:clusterNYC}
    \end{subfigure}
    \begin{subfigure}[b]{0.4\columnwidth}
         \centering
         \includegraphics[width=\textwidth]{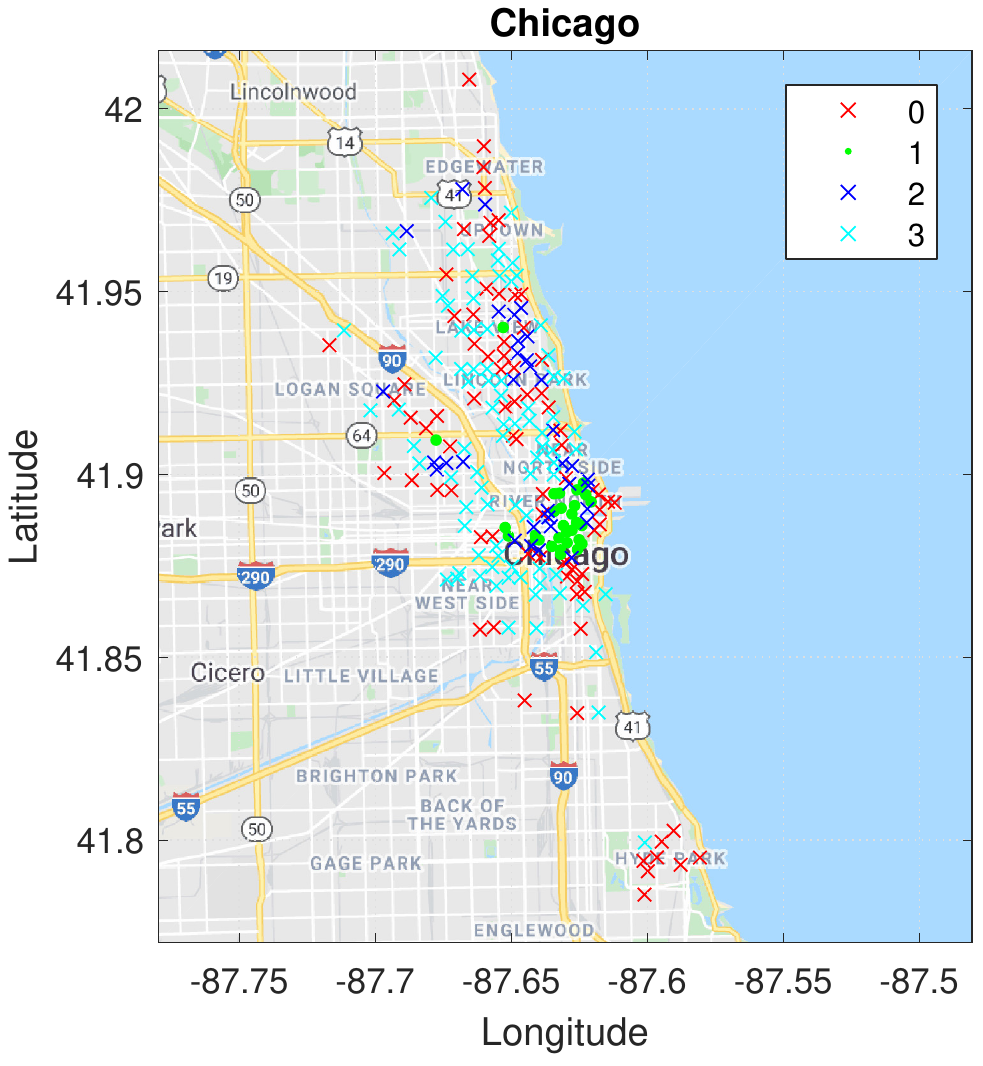}
         \label{fig:clusterChicago}
    \end{subfigure}
    \vspace{-0.2in}
    \caption{
    The cluster results of existing and new stations. The clusters represented by dots contain new stations, while the cross ones have only existing stations. 
    }
    \label{fig:cluster}
\end{figure}

We adopt the K-means clustering algorithm to find out the clusters from the
normalized station-centered heatmaps constructed as in 
Section \ref{sec3.3}. 
We first calculate the sum of all the $\mathcal{G}_{lat}\times\mathcal{G}_{lon}$ entries at each channel, and the mean of heatmaps over the timeframe $[T, T']$, generating vectors of length $\mathcal{P}$. The POI metric scores between the heatmaps of stations $j$ and $k$ for the clustering 
are calculated from the Euclidean distance between the generated vector: 
\begin{equation}
Sim_{j,k} = \left\|\sum_{e = 1}^{\mathcal{G}_{lat}\times\mathcal{G}_{lon}}\overline{\mathbf{H}}_{j}^{\normalfont {\textit{e}}} - \sum_{e = 1}^{\mathcal{G}_{lat}\times\mathcal{G}_{lon}}\overline{\mathbf{H}}_{k}^{\normalfont {\textit{e}}}\right\|.
\end{equation}
The combination of both existing and new stations
in each city are then clustered based on the above metric scores. 
The results of clustering existing and new stations
for NYC and Chicago 
based on {$11\times11\times\mathcal{P}$} heatmaps are shown below on Figure \ref{fig:cluster}.

\section{\name{}: Usage Prediction for Station Network Reconfiguration}\label{sec:model}
We further discuss the details of \name{} model  in this section. 
We first define our usage problem,
followed by a description of model structure in Section \ref{model}. 
Our model is trained on the datasets of active existing stations, and the details will be discussed in Section \ref{training}.

\subsection{Problem \& Model Definitions}\label{model}
\subsubsection{Problem Definition}
The problem in this study can be defined as:
given the input station-centered heatmaps, $\mathbf{H_{n,\tau'}}$,  and the external features $\mathbf{ex_{\tau'}}$, including the weather conditions and weekend/holidays information, 
for each station $n$ at the period of time $\tau' \in [1,\mathcal{T}]$, either an existing or new one after reconfiguration,
predict future usage (pick-ups and drop-offs), $L_{n,\tau}$, where $\tau = \mathcal{T} + 1$ is 
the target time interval for an existing station or the one when the new station is established after reconfiguration. 

\subsubsection{Model Overview}

\begin{figure}
    \centering
    \includegraphics[width=\columnwidth]{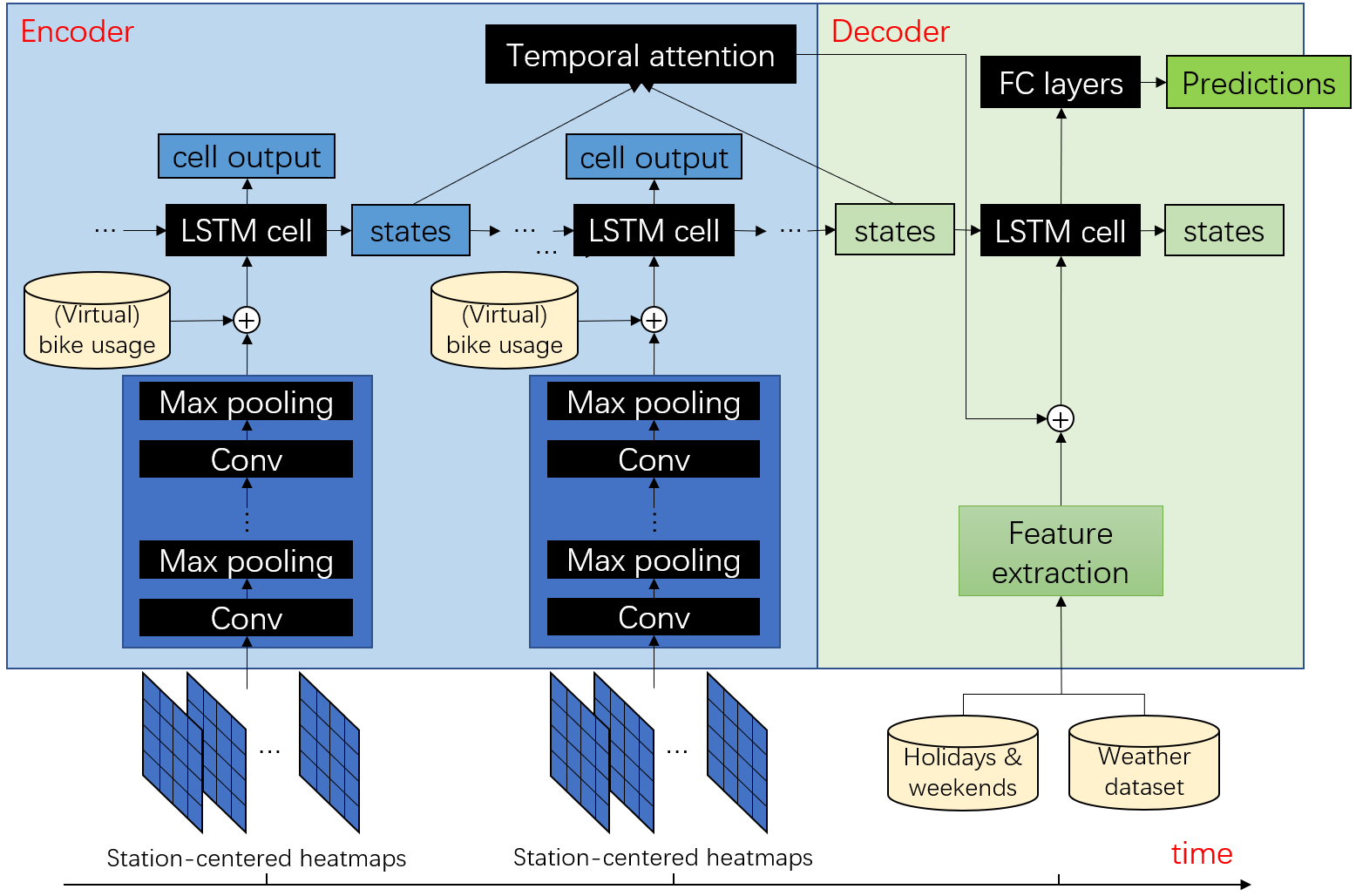}
    \caption{Illustration of model designs in \name{}.}
    \label{fig:model}
\end{figure}

Our model consists of three major modules. 
\begin{enumerate}
\item[1)] \textit{Spatial feature learning}: A deep-channel \texttt{CNN} learns the station specific heatmaps which describe the driving force of usage for each particular station based on its spatial characteristics. 
\item[2)] \textit{Temporal feature learning}: The output of \texttt{CNN} combined with historical usage data is then fed as input of a \texttt{LSTM} which learns the temporal pattern of station's usage. 
\item[3)] \textit{Feature attention characterization}:
A temporal attention mechanism is applied to further capture and differentiate the correlations between features across different timestamps.
\end{enumerate}
 An overview of \name{} is further illustrated in Figure \ref{fig:model}. The details of each module will be presented as follows.

\subsubsection{Convolutional Neural Network (\texttt{CNN})} 
As described before, the station-specific or station-centered heatmaps as the representation of spatial features, i.e. the POIs distribution and regional usage popularity, indicate the motivation of people heading to places around and hence correlate with the bike usage. 
Differences in this driving force may result in different directions of bike usage and are represented by the designed heatmaps. However, the mechanism behind this driving force of bike usage can be too complicated to formulate, and simple vector concatenation is not enough. Therefore, we propose a deep-channel \texttt{CNN} model to learn the correlation between station specific spatial features in an area and station bike usage. Note that different from previous studies~\cite{ma2017learning}, we focus on learning station-centered heatmaps to further identify the useful correlations between the center station's usage and the neighborhood.

The \texttt{CNN} has $C$ convolutional layers, each of which is followed by a max pooling layers. The size of the convolutional layers are $h \times w \times c$, all with \texttt{relu} activation function. A fully connected layer is implemented as the last layer to generate the output of \texttt{CNN} of size 1.

\subsubsection{Long Short-Term Memory (LSTM)}
The output of the \texttt{CNN} is concatenated with historical usage data as the input of \texttt{LSTM}. The last hidden state of the \texttt{LSTM} cell is connected to a fully connected layer to generate predictions. The equations of an \texttt{LSTM} cell at timestamp $t$ are listed below:
\begin{equation}
\begin{aligned}
    \mathbf{i}_{t}&=\sigma\left(\mathbf{W}^{i}\left[\mathbf{h}_{t-1},\mathbf{x}_{t}\right]+\mathbf{b}^{i}\right), 
    \\
    \mathbf{f}_{t}&=\sigma\left(\mathbf{W}^{f}\left[\mathbf{h}_{t-1},\mathbf{x}_{t}\right]+\mathbf{b}^{f}\right),
    \\
    \mathbf{o}_{t}&=\sigma\left(\mathbf{W}^{o}\left[\mathbf{h}_{t-1},\mathbf{x}_{t}\right]+\mathbf{b}^{o}\right),
    \\
    \mathbf{\tilde{c}}_{t}&=\text{tanh}\left(\mathbf{W}^{i}\left[\mathbf{h}_{t-1},\mathbf{x}_{t}\right]+\mathbf{b}^{i}\right),
    \\
    \mathbf{c}_{t} &= \left(\mathbf{f}_{t}*\mathbf{c}_{t-1}+\mathbf{i}_{t}*\mathbf{\tilde{c}}_{t}\right),
    \\
    \mathbf{h}_{t} &=\text{tanh}\left(\mathbf{c}_{t}\right)*\mathbf{o}_{t},
\label{LSTM}
\end{aligned}
\end{equation}
where $\mathbf{x}_{t}\in \mathbb{R} ^{l}$ is the input of this timestamp of size $l$, which is the concatenation of historical usage and the \texttt{CNN} output, $\mathbf{c}_{t}$ and $\mathbf{h}_{t}$ are the cell state and hidden state of \texttt{LSTM} cell at time $t$, $\mathbf{i}_{t}$, $\mathbf{f}_{t}$, $\mathbf{o}_{t}$ and $\mathbf{\tilde{c}}_{t}$ are intermediate variables of \texttt{LSTM} cell, and $\mathbf{W}^{q} \in \mathbb{R} ^{d \times (d+l)}$ and $\mathbf{b}^{q} \in \mathbb{R} ^{d}$ where $q=\mathbf{i}, \mathbf{f}, \mathbf{o}$ are trainable variables, and $d$ is the number of hidden units of \texttt{LSTM} cell.

\subsubsection{Temporal Attention}
The temporal attention mechanism captures correlation between features across different timestamps. The intuition behind this is that there is a strong temporal dependency of the usage of a station at time $t$, for example, on the usage at the same time one day ago. The temporal attention scores between the current decoder \texttt{LSTM} timestep $t$ and one of the previous encoder \texttt{LSTM} hidden states, $\lambda_{t,t'} $, are calculated as Eq. (\ref{eq_tmps}) by a concatenation manner:
\begin{equation}
\lambda_{t,t'} = \mathbf{v}_{a}^\intercal \texttt{tanh}\left(\mathbf{W}_{a}[\mathbf{h}_{t-1};\mathbf{c}_{t-1}] + \mathbf{U}_{a}\mathbf{h}_{t'} + \mathbf{b}_{a}\right),
\label{eq_tmps} 
\end{equation}
where $\mathbf{v}_{a}, \mathbf{b}_{a} \in \mathbb{R} ^{d}$, $\mathbf{W}_{a} \in \mathbb{R} ^{d \times 2d}$ and $\mathbf{U}_{a} \in \mathbb{R} ^{d \times d}$ are learnable parameters, $d$ is the number of hidden units of \texttt{LSTM} cell, $\mathbf{h}_{t}$ and $\mathbf{c}_{t}$ are hidden state and cell state of decoder at timestamp $t$, and $\mathbf{h}_{t'}$ is hidden state of encoder at timestamp $t'$, which is in range of $[1,\mathcal{T}]$. 
The attention weight, denoted as $\gamma_{t,t'}$, is then a \texttt{softmax} function of $\lambda_{t,t'} $ as in Eq. (\ref{eq_tmpw}):
\begin{equation}
\gamma_{t,t'} = \frac{\texttt{exp}\left(\lambda_{t,t'}\right)}{\sum_{t = 1}^{\mathcal{T}}\texttt{exp}\left(\lambda_{t,t'}\right)}.
\label{eq_tmpw} 
\end{equation}
The weighted sum of the encoder \texttt{LSTM} hidden states, $\mathbf{d}_{t}$, shown in Eq. (\ref{eq_tmpo}) concatenated with external features $\mathbf{ex_{t}}$, $[\mathbf{d}_{t}:\mathbf{ex_{t}}]$, serves as the input of decoder:
\begin{equation}
\mathbf{d}_{t}=\sum_{t'=1}^{\mathcal{T}}\gamma_{t,t'} \mathbf{h}_{t'},
\label{eq_tmpo} 
\end{equation}
where $\mathcal{T}$ is the number of encoder timestamps.

\subsection{Model Training \& Prediction for New Stations}\label{training}
As conventional practices, we handle the existing station prediction based on the historical data available. 
Given absence of historical data, the predictions of the usage for new stations are based on the model trained on the active existing stations. To address the initialization problem of the model, 
during the training process, we generate the virtual trip data based on the weighted average of those from multiple existing stations.
This is achieved by randomly picking a batch from the samples of
all the existing stations at each training epoch. 
This way, the model is able to learn their shared bike usage patterns across all the existing stations, and such knowledge learned can be used to predict the usage of new stations, considering that
the new stations share correlated spatial and temporal usage features
of existing stations.

The model is trained upon the existing stations within each cluster. 
To address the model initialization without historical data, for the new stations we design an efficient mechanism to generate
the virtual historical usage from those of surrounding existing stations. We note that the two adjacent stations have similar bike usage and mobility patterns because of their similar spatial-temporal characteristics. Therefore, we generate the virtual bike usage by the distances between stations.
Specifically, we compute the geographic similarity score between a new station, $n$, and an existing peer in its neighborhood, $f$, based on their mutual geographic distance in km:
\begin{equation}
Sim_{n,f}= \frac{1}{\text{distance}\left(n,f\right)}. 
\label{sim} 
\end{equation}
Then the similarity scores across all the existing stations is normalized 
to find the weights for the known bike usage assigned upon each existing station:
\begin{equation}
\omega_{n,f} = \frac{Sim_{n,f}^{2}}{\sum_{f = 1}^{\mathbf{F}}Sim_{n,f}^{2}}.
\label{sfm} 
\end{equation}

The virtual bike usage for the new station at time $\tau$ based on the historical usage of those existing stations, $L_{f,\tau}$ is finally calculated by:
\begin{equation}
L_{n,\tau}=\sum_{f\in\mathbf{F}}^{\mathbf{F}}\omega_{n,f} L_{f,\tau},
\label{simulate_new} 
\end{equation}
where $\mathbf{F}$ is the set of existing stations.

\section{Experimental Studies}\label{sec:experiment}

In this section, we present the experimental results
based on the datasets of Citi (NYC), Divvy (Chicago) and Metro (LA). The experimental settings are first introduced in Section \ref{exp_setting}, followed by experimental results on existing/new stations in Section \ref{results}. 

\subsection{Experimental Settings}\label{exp_setting}

We compare \name{} with the following baselines and state-of-art models on the given datasets:
\begin{enumerate}
    \item \textit{ARIMA}: Auto Regressive Integrated Moving Average for time-series forecasting. The size of sliding window is set to be 24.
    \item \textit{RNN}: Simple Recurrent Neural Network for time-series predictions \cite{pan2019predicting}. The length of input sequence is 24.
    \item \textit{LSTM}: Long Short-Term Memory neural network \cite{lin1996learning} predicts future usage with historical data of last 24 hour.
    \item \textit{GRU}: Gated Recurrent Units as another recurrent neural network for time-series predictions \cite{cho2014properties}.
    \item \textit{GCN}: Graph Convolutional Neural Networks with Data-driven Graph Filter \cite{chai2018bike} predicts station-level usage taking into account of station correlations.
\end{enumerate}

We are leveraging the features for previous 24 hours to predict the station-level bike usage for the next timestamp.
For the evaluation of the existing stations for NYC and Chicago datasets, our model as well as comparison schemes are trained based on the hourly usage data from April 11, 2019 to July 19, 2019 (2,400 hours duration), and are tested on the hourly usage data from July 20, 2019 to August 18, 2019 (720 hours duration). 
For LA datasets, the models are trained based on the 4-hour bike usage data in June 2019 (720 hours duration), and are tested on the data for the following 30 days (720 hours duration).

For the evaluation of the new stations, we find out all the deployed stations 
from May to August 2019 for NYC and Chicago, and those from June to December 2019 for LA.
Our model as well as the baselines are tested on the usage data within 4 weeks (672 hours duration) 
for NYC and Chicago and 2 weeks for LA
from the first appearance of the hourly usage of the stations. 
The stations with sufficient bike usage are of our interest here, and we discard those with little usage (say, less than 10 pick-ups/drop-offs per day). 
For new stations in LA, due to
the large sparsity in the bike usage data, we consider the appearance of consecutive bike usage as the first usage and the start of our evaluations. 
The numbers of existing/new stations studied in our experimental evaluations are further identified and listed as follows. 
The numbers of existing and new stations studied, (existing, new), for each city are as follows:
(631, 15) for NYC,
(229, 13) for Chicago, (48, 7) for LA.
The existing and new stations are clustered together in the manner described in Section \ref{seccluster}.
We train \name{} and baselines models on all the clusters of the three bike sharing systems. Then we evaluate models regarding their predictions of new stations.

\begin{figure}
    \centering
         \includegraphics[width=\columnwidth]{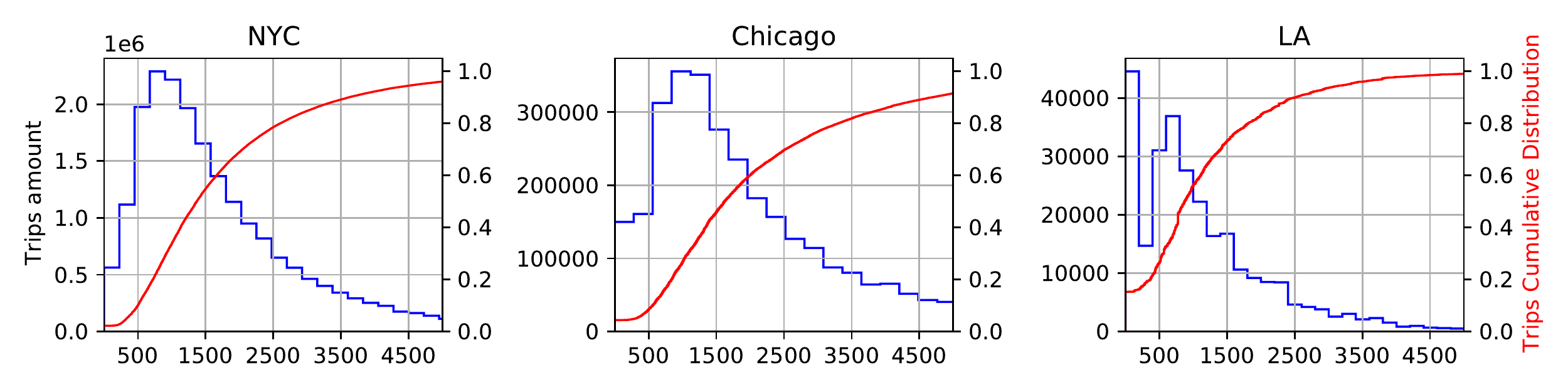}
    \caption{Distributions of bike trip distance (meters) for NYC, Chicago and LA.} %
    \label{fig:distdist}
    \vspace{-0.1in}
\end{figure}

Figure \ref{fig:distdist} shows that a majority of bikes rented at one station in 
NYC, Chicago and LA, are returned at stations beyond 500m away, demonstrating that for a specific station there is a small number ($<20\%$) of bike transitions between the stations within a grid of 500m $\times$ 500m and the bikes rented at this region are mostly returned to regions somewhere else. 
In addition, a majority of bike trips are within the range of 500m and 5,000m as shown in Figure \ref{fig:distdist}. 
Therefore, the size of station-centered heatmaps discussed in Section \ref{heatmaps} is set to be $11\times11\times\mathcal{P}$, where $\mathcal{P}$ is 2 (regional pick-ups and drop-offs) plus the total number of POIs categories  (13 for NYC, 7 for Chicago and 15 for LA), and each grid of the heatmaps is a 500m $\times$ 500m area on the city map.
This way, the heatmaps cover most of the areas where the riders can reach by the bikes rented from the center station. 

Other model parameters of \name{} are set as followed.
The \texttt{CNN} component has 3 layers convolutional layers with sizes  $h \times w \times c$; we set $3 \times 3 \times 256$, $3\times3\times128$ and $2\times2\times64$, respectively, and \texttt{relu} activation function for each of them. Each of the first two convolutional layers is followed by a maxpooling layer. 
The final convolutional layer is connected with a fully connected layer which converts the
\texttt{CNN}'s output into the one of size 1. The number of layers of \texttt{LSTM} is 1,024. The dropout rate for the \texttt{LSTM} is 0.5. The learning rate is 0.001. The batch size is set to be 128. Total number of training epochs is 5,000. We have implemented \name{} and other schemes in Python 3.7 and Tensorflow 2.1, and the models are trained and evaluated upon a desktop server with Intel i5-8700K, 16GB RAM, Nvidia GTX 1060/1050Ti and Windows 10.

We use mean square error (\textit{MSE}) as the training metric, 
and we evaluate model performance as well as the results based on both \textit{MSE} and mean abosolute error (\textit{MAE}):
\begin{equation}
\begin{aligned}
    MSE = \frac{1}{M}\sum_{i}^{M}\left(y_{i}-\hat{y}_{i}\right)^{2},\quad
    MAE = \frac{1}{M}\sum_{i}^{M}\left|y_{i}-\hat{y}_{i}\right|,
\label{metrics}
\end{aligned}
\end{equation}
where \textit{M}, $y_{i}$ and $\hat{y}_{i}$ are the total number of predictions made by the evaluated model, the ground-truth of the bike usage and the predicted bike usage, respectively.

\subsection{Experimental Results}\label{results}

\subsubsection{Usage Prediction for Existing Stations}
First, we test our model on all active existing stations for all three datasets. 
The accuracy of bike usage predictions across the existing stations is shown in Table \ref{tab:fixedcom}.
It is shown that our model achieves overall better accuracy compared with other baselines for the systems in NYC, Chicago and LA.
It is mainly because the station-centered heatmaps incorporate the trend that the bike riders travel between the stations. 
Focusing upon each station's neighborhood, a heatmap characterizes the mobility patterns, which are learned and captured by \name{}, and thus the bike usage at the new stations with similar neighborhood patterns can be further predicted. 
For Metro in LA, \texttt{GCN} achieves comparable or slightly better performance like \name{}, likely due to the sparsity of bike station network in LA. 
Despite this, 
\name{} has shown high accuracy and robustness
with complicated mobility patterns.

\begin{figure*}
    \centering
    \begin{subfigure}[b]{0.325\textwidth}
         \centering
         \includegraphics[width=\textwidth]{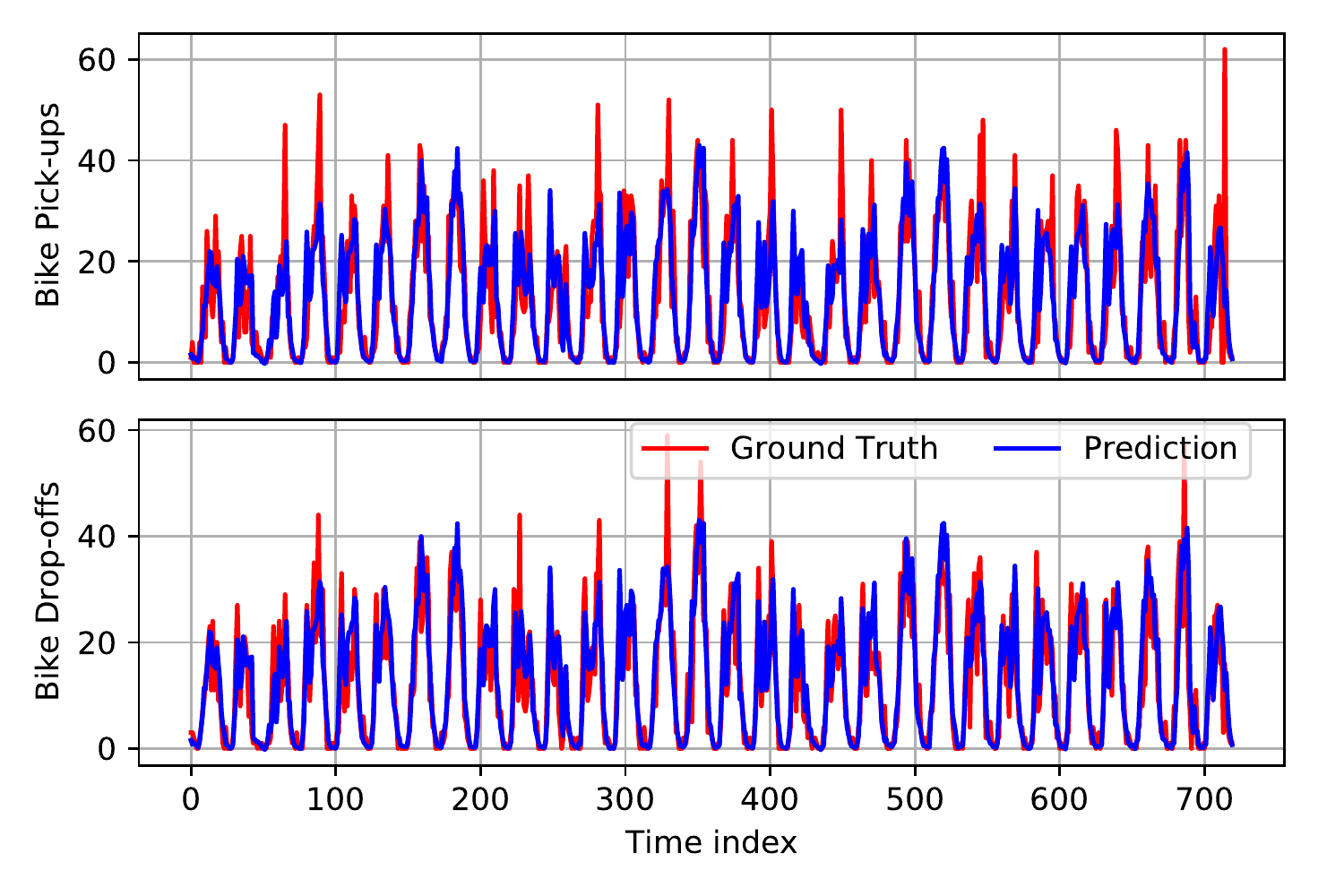}
         \vspace{-0.1in}
         \caption{A station at NYC (40.7659$^\circ$, -73.9763$^\circ$). }
    \end{subfigure}
    \begin{subfigure}[b]{0.325\textwidth}
         \centering
    \includegraphics[width =\textwidth]{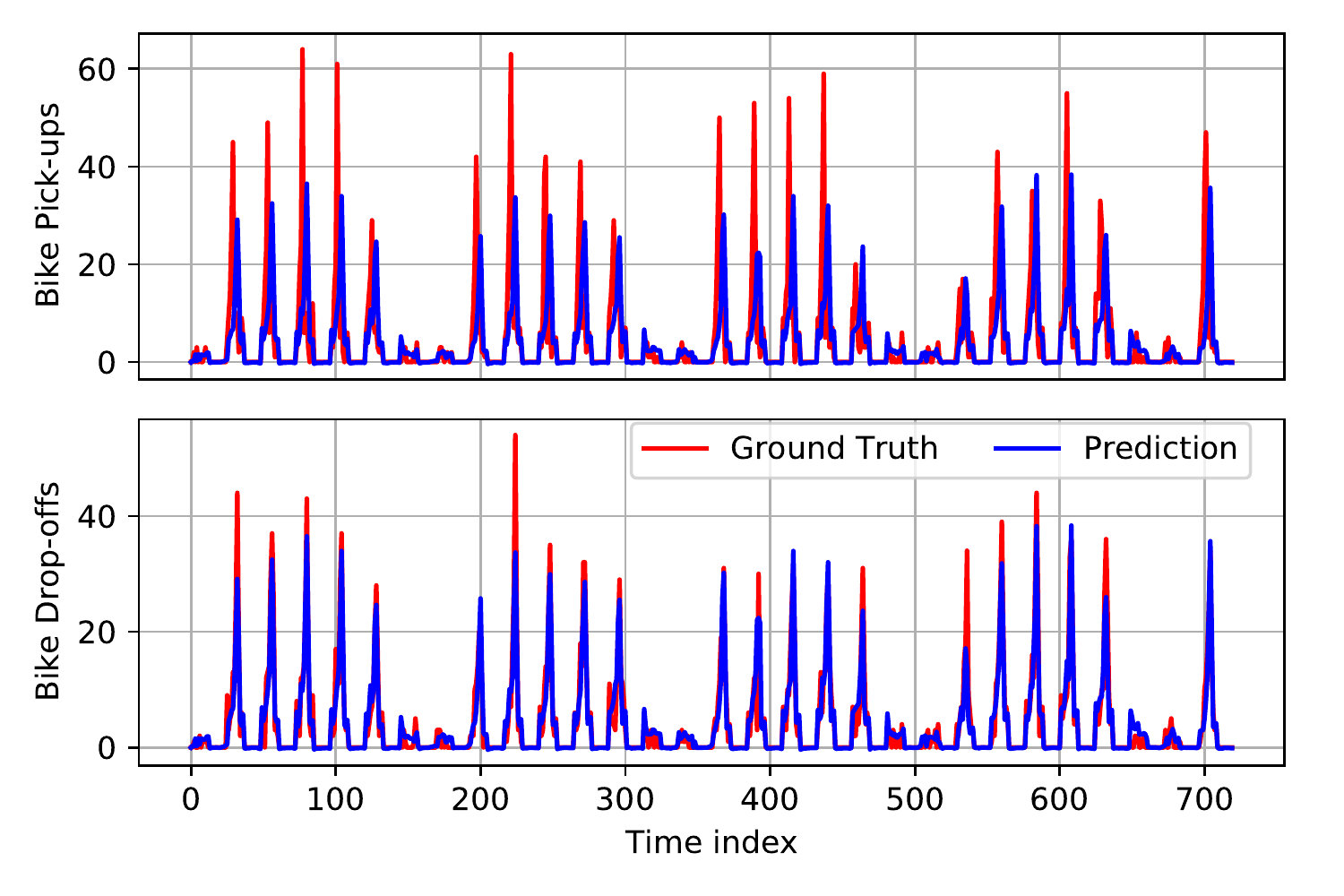}
    \vspace{-0.1in}
         \caption{A station at Chicago (41.8782$^\circ$, -87.6319$^\circ$). }
    \end{subfigure}
    \begin{subfigure}[b]{0.325\textwidth}
         \centering
    \includegraphics[width =\textwidth]{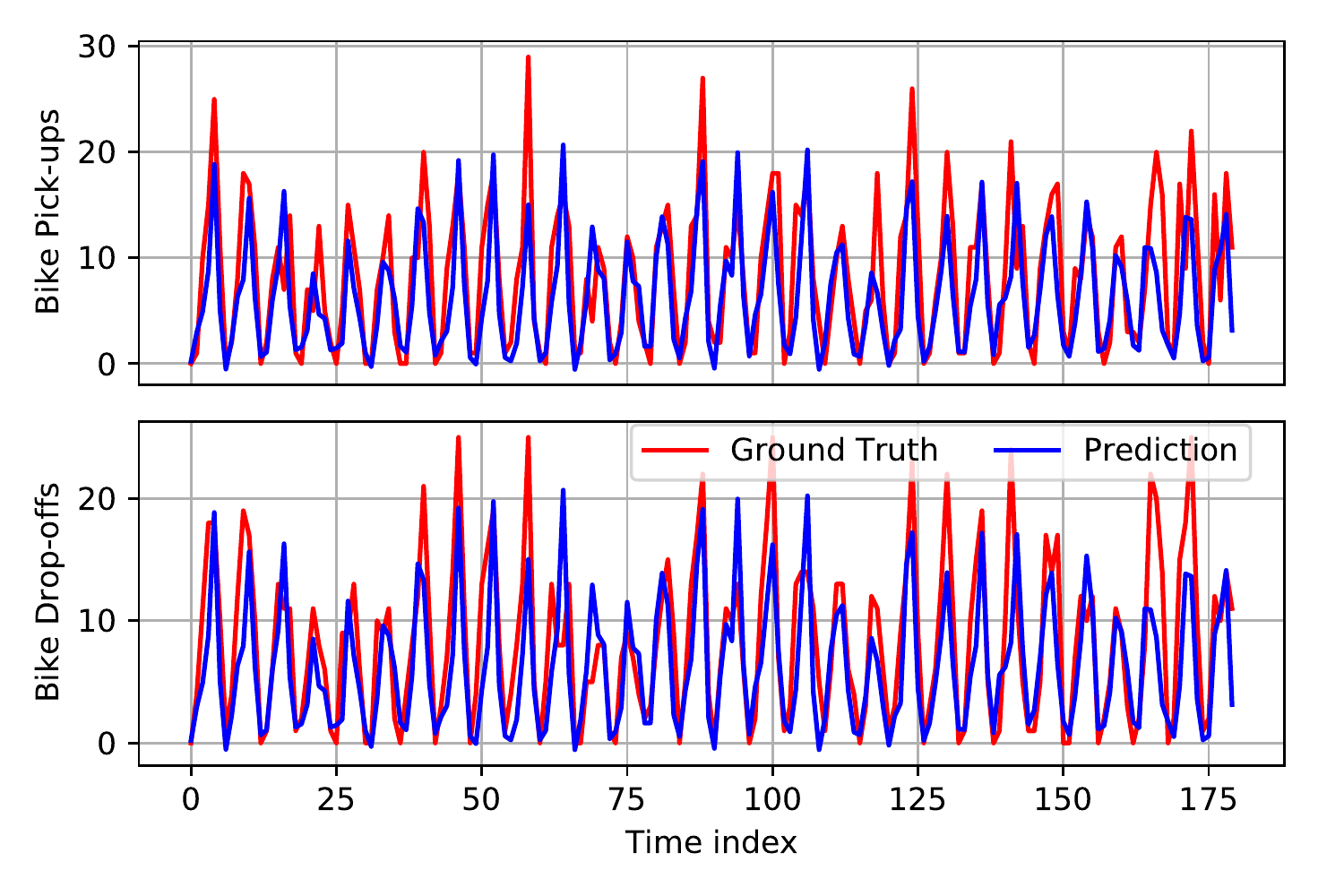}
    \vspace{-0.1in}
         \caption{A station at LA (34.085$^\circ$,-118.259$^\circ$). }
    \end{subfigure}
    \vspace{0.1in}
    \caption{Hourly bike usage prediction for the existing stations in the three datasets.}
    \label{figfixed}
\end{figure*}

Figure \ref{figfixed} illustrates the bike usage (pick-ups and drop-offs) predictions for the three existing stations in NYC, Chicago and LA. We can observe that the predictions of \name{} are highly accurate, and close to the ground-truth measurements. Such accuracy can enable advanced bike station operational applications like station demand and supply rebalancing.

\begin{table}[]
\caption{Comparison between \name{} and other baseline and state-of-art models for existing stations of the three datasets.}
\label{tab:fixedcom}
\resizebox{\columnwidth}{!}{%
\begin{tabular}{|c|c|c|c|c|c|c|c||c|}
\hline
\multicolumn{3}{|c|}{\textbf{Schemes}}                                              & \texttt{ARIMA}  & \texttt{RNN}    & \texttt{GCN}    & \texttt{LSTM}   & \texttt{GRU}    & \textbf{\name{}}  \\ \hline
\multirow{4}{*}{\textbf{Citi}}  & \multirow{2}{*}{\textbf{Pick-up}}  & \textbf{MAE} & 2.682  & 2.345  & 2.240  & 2.164  & 2.166  & 1.771  \\ \cline{3-9} 
                                &                                    & \textbf{MSE} & 30.028 & 18.711 & 12.226 & 16.080 & 16.143 & 10.599 \\ \cline{2-9} 
                                & \multirow{2}{*}{\textbf{Drop-off}} & \textbf{MAE} & 2.564  & 2.265  & 2.084  & 2.100  & 2.112  & 1.730  \\ \cline{3-9} 
                                &                                    & \textbf{MSE} & 24.762 & 16.514 & 11.318 & 14.210 & 14.245 & 9.447  \\ \hline
\multirow{4}{*}{\textbf{Divvy}} & \multirow{2}{*}{\textbf{Pick-up}}  & \textbf{MAE} & 1.964  & 1.832  & 2.300  & 1.437  & 1.418  & 1.263  \\ \cline{3-9} 
                                &                                    & \textbf{MSE} & 23.693 & 14.866 & 6.382  & 11.293 & 10.732 & 7.687  \\ \cline{2-9} 
                                & \multirow{2}{*}{\textbf{Drop-off}} & \textbf{MAE} & 1.927  & 1.633  & 1.067  & 1.247  & 1.404  & 1.247  \\ \cline{3-9} 
                                &                                    & \textbf{MSE} & 21.075 & 13.584 & 6.382  & 10.937 & 10.508 & 7.980  \\ \hline
\multirow{4}{*}{\textbf{Metro}} & \multirow{2}{*}{\textbf{Pick-up}}  & \textbf{MAE} & 2.126  & 3.258  & 1.276  & 1.878  & 1.936  & 1.462  \\ \cline{3-9} 
                                &                                    & \textbf{MSE} & 8.578  & 16.071 & 4.895  & 7.471  & 7.26   & 4.879  \\ \cline{2-9} 
                                & \multirow{2}{*}{\textbf{Drop-off}} & \textbf{MAE} & 2.154  & 1.478  & 1.480  & 1.251  & 1.005  & 1.435  \\ \cline{3-9} 
                                &                                    & \textbf{MSE} & 8.752  & 4.535  & 3.965  & 3.393  & 2.245  & 4.610  \\ \hline
\end{tabular}
}
\end{table}

\begin{table}[]
\caption{Performance comparison
for predictions of new stations. }
\label{tab:new}
\resizebox{0.8\columnwidth}{!}{%
\begin{tabular}{|c|c|c|c|c|c||c|}
\hline
\multicolumn{3}{|c|}{\textbf{Schemes}}                                              & \texttt{RNN}    & \texttt{LSTM}   & \texttt{GRU}    & \textbf{\name{}} \\ \hline
\multirow{4}{*}{\textbf{Citi}}  & \multirow{2}{*}{\textbf{Pick-up}}  & \textbf{MAE} &   2.922     &      2.760  &    2.736    & 2.495                \\ \cline{3-7} 
                                &                                    & \textbf{MSE} &   20.961     &    18.652    & 18.461       & 15.883               \\ \cline{2-7} 
                                & \multirow{2}{*}{\textbf{Drop-off}} & \textbf{MAE} & 2.774  & 2.728  & 2.738  & 2.568                \\ \cline{3-7} 
                                &                                    & \textbf{MSE} & 17.833 & 18.930 & 19.016 & 15.835               \\ \hline
\multirow{4}{*}{\textbf{Divvy}} & \multirow{2}{*}{\textbf{Pick-up}}  & \textbf{MAE} &   2.092     &     1.814   &  1.825      & 1.627                \\ \cline{3-7} 
                                &                                    & \textbf{MSE} &    23.475    &    21.089    &     21.025   & 13.282               \\ \cline{2-7} 
                                & \multirow{2}{*}{\textbf{Drop-off}} & \textbf{MAE} & 1.812  & 1.589  & 1.639  & 1.442                \\ \cline{3-7} 
                                &                                    & \textbf{MSE} & 14.126 & 11.491 & 12.317 & 8.223                \\ \hline
\multirow{4}{*}{\textbf{Metro}} & \multirow{2}{*}{\textbf{Pick-up}}  & \textbf{MAE} &    1.552    &     1.249   &   1.246     &          1.098          \\ \cline{3-7} 
                                &                                    & \textbf{MSE} &   3.286     &     3.425   &  3.135      &               2.439       \\ \cline{2-7} 
                                & \multirow{2}{*}{\textbf{Drop-off}} & \textbf{MAE} &   1.588     &    1.583    &    1.216    &           1.043           \\ \cline{3-7} 
                                &                                    & \textbf{MSE} &   3.683     &      3.662  &    2.463    &           2.795          \\ \hline
\end{tabular}
}
\end{table}

\begin{table}[]
\centering
\caption{Comparison of performance between \name{} w/ and w/o virtual generated data.}
\label{tab:newvirtual}
\resizebox{\columnwidth}{!}{%
\begin{tabular}{|c|l|c|c|c|c|c|c|}
\hline
\multicolumn{2}{|c|}{\multirow{2}{*}{\textbf{Schemes}}}                            & \multicolumn{2}{c|}{\textbf{Citi}} & \multicolumn{2}{c|}{\textbf{Divvy}} & \multicolumn{2}{c|}{\textbf{Metro}} \\ \cline{3-8} 
\multicolumn{2}{|c|}{}             & \textbf{MAE}     & \textbf{MSE}    & \textbf{MAE}     & \textbf{MSE}     & \textbf{MAE}     & \textbf{MSE}     \\ \hline
\multirow{2}{*}{\textbf{Pick-up}}  & \textbf{\name{} w/ virtual}  & 2.384            & 15.018          & 1.638            & 11.983           &       1.305           &        4.639          \\ \cline{2-8} 
                                   & \textbf{\name{} w/o virtual} & 2.387            & 16.661          & 1.685            & 14.528           &          1.395        &           5.014       \\ \hline
\multirow{2}{*}{\textbf{Drop-off}} & \textbf{\name{} w/ virtual}  & 2.666            & 15.892          & 1.589            & 9.320            &      1.561            &       5.797           \\ \cline{2-8} 
                                   & \textbf{\name{} w/o virtual} & 2.547            & 16.510          & 1.522            & 10.300           &      1.727            &       7.929           \\ \hline
\end{tabular}
}
\end{table}

\subsubsection{Bike Usage Prediction for New Stations} \label{sec:expnew}
The predictions of bike usage for new stations can be done using the knowledge learned from the existing stations. 
Given the trained models from all the existing stations, \texttt{RNN}, \texttt{LSTM} and \texttt{GRU} can be directly applied to predict each individual new stations, and therefore we focus on comparing \name{} with the three approaches here. 
We leverage the knowledge learned from the existing stations for the predictions of new stations given the bike station network reconfiguration. 
MAE and MSE of new stations' usage predictions, pick-ups and drop-offs, by \name{} as well as three baselines, \texttt{RNN}, \texttt{LSTM} and \texttt{GRU}, are shown in Table \ref{tab:new}. As is shown, \name{} outperforms those three baseline models in predicting the usage of new stations for all three bike sharing systems.

We illustrate the performance with and without the virtual bike usage data generated. 
Virtual bike usage one day ahead of the first bike usage of new stations are generated as
in Section \ref{training} as the starting points.
The virtual usage provides an initial inference for the models regarding how the mobility pattern of a new station possibly look like, enhancing the model accuracy 
as is shown in Table \ref{tab:newvirtual}. Here we only compare the MAE and MSE for predicting the next 24 hours after the first use of the stations given the data availability.

\section{Discussion}\label{sec:discussion}

\textit{Incorporating Other Information}: Despite the features selected in this work as described in Section \ref{sec3.1}, other features may also be correlated with bike usage, such as events and demographic distributions \cite{chen2016dynamic}. However, such information is not considered due to the limit of our current resources. Our prediction accuracy could be further increased with the inclusion of those features in our model. Nevertheless, the generic design of station-centered heatmaps as the feature representations in this study allows the easy integration of other information.  

\noindent\textit{Sparsity of Usage Data}: 
Though we focus on active existing stations in this study, a lot of stations are not so active %
that their bike usage is low, especially in LA %
. For experimental study of LA bike usage, we consider 4-hour time interval as one timestamp to lower the influence of data sparsity. Predicting the bike usage with sparsity is challenging, yet important for system management. Further study on how to deal with sparse historical data is necessary.  

\noindent
\textit{Similarity Scores for Virtual Data Generation}:
As mentioned in Section \ref{sec:expnew}, virtual historical usage
is essential for the first few predictions, and their accuracy depends on the quality of the initial inference
in addition to the model performance. 
In this work, we compute the virtual historical usage based on the mutual geometric distances between stations, 
which has been shown to benefit predicting the
new stations. 
However, further enhanced inference can 
be achieved by inclusion of other information such as crowd's awareness of new stations, where a comprehensive mechanism is needed for future study.

\section{Conclusion}\label{sec:conclusion}
In this work, we propose a novel bike station prediction algorithm called \name{} for predicting station-based bike usage of future stations given bike station network reconfiguration.
We design novel station-centered
heatmaps which characterize for each target station centered at the heatmap
the trend that the riders travel between it and the neighboring regions, making the common patterns of the bike station network learnable. \name{} further leverages such knowledge learned to predict the usage for new stations with the aid of virtual historical usage of new stations generated according to their correlation to the surrounding existing stations.
Extensive experiment study on the bike sharing systems of NYC, Chicago and LA shows \name{} is capable in predicting usage at new/future stations as well as existing stations, and outperforms the baseline and state-of-art models in our experimental evaluations.

\newpage
\bibliographystyle{ACM-Reference-Format}
\bibliography{references}

\end{document}